\documentclass{article}

\usepackage[preprint]{neurips_2024}

\usepackage[pdftex]{graphicx}
\usepackage[dvipsnames]{xcolor}
\usepackage[utf8]{inputenc} % allow utf-8 input
\usepackage[T1]{fontenc}    % use 8-bit T1 fonts
\usepackage{hyperref}       % hyperlinks
\usepackage{url}            % simple URL typesetting
\usepackage{booktabs}       % professional-quality tables
\usepackage{amsfonts}       % blackboard math symbols
\usepackage{nicefrac}       % compact symbols for 1/2, etc.
\usepackage{microtype}      % microtypography
\usepackage{xcolor}         % colors
\usepackage{float}
\usepackage[raster,most]{tcolorbox}
\usepackage{natbib}
\setcitestyle{numbers,square}

\title{R1-Zero’s “Aha Moment” in Visual Reasoning on a 2B Non-SFT Model}

\author{%
Hengguang Zhou\textsuperscript{*} \\
University of California, LA\\
\And 
Xirui Li\textsuperscript{*} \\
University of California, LA\\
\AND Ruochen Wang\textsuperscript{$\dagger$} \\
University of California, LA \\
\AND Minhao Cheng \\
Pennsylvania State University\\
\And Tianyi Zhou  \\
University of Maryland\\
\And Cho-Jui Hsieh \\
University of California, LA\\
\And $*$ Equal contribution \quad $\dagger$ Main Advisor \\
}

\begin{document}

\maketitle

\begin{center}
    \vspace{-5mm}\includegraphics[width=0.2\textwidth]{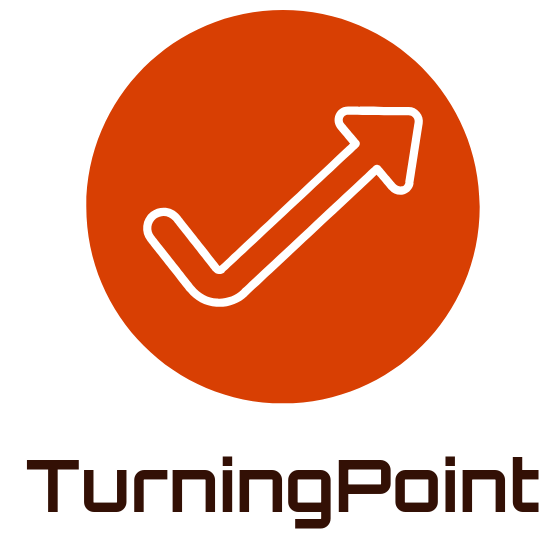}
\end{center}
\begin{abstract}
The recent DeepSeek-R1 demonstrated how reinforcement learning with simple rule-based reward can enable autonomous development of complex reasoning in large language models, characterized by the "aha moment", in which the model manifest self-reflection and increased response length during training. However, attempts to extend this success to multimodal reasoning often failed to reproduce these key characteristics.
In this report, we present the first successful replication of these emergent characteristics for multimodal reasoning on only a non-SFT 2B model. Starting with Qwen2-VL-2B and applying reinforcement learning directly on the SAT dataset, our model achieves \textbf{59.47\%} accuracy on CVBench, outperforming the base model by approximately \textbf{\textasciitilde30\%} and exceeding both SFT setting by \textbf{\textasciitilde2\%}. In addition, we share our failed attempts and insights in attempting to achieve R1-like reasoning using RL with instruct models, aiming to shed light on the challenges involved. Our key observations include: (1) applying RL on instruct model often results in trivial reasoning trajectories, and (2) naive length reward are ineffective in eliciting reasoning capabilities. The project code is available at \href{https://github.com/turningpoint-ai/VisualThinker-R1-Zero}{https://github.com/turningpoint-ai/VisualThinker-R1-Zero}

\end{abstract}
\definecolor{MyGreen}{cmyk}{0.32, 0.0, 0.06, 0.24}
\begin{tcolorbox}[title=Work in Progress, label=WIP statement,  width=\textwidth,center, colbacktitle=MyGreen, colback=White]
This research is still a work in progress. In this report, we share our successes, failures, and insights from our preliminary exploration of potential approaches for developing R1-like reasoning in multimodal models. This is an ongoing effort and we will be committed to continuous update and improvement. This report presents only a minimal version, with the full version to be released soon.
\end{tcolorbox}

\section{Introduction}
\begin{figure}[h!]
    \centering
    \includegraphics[width=1\textwidth]{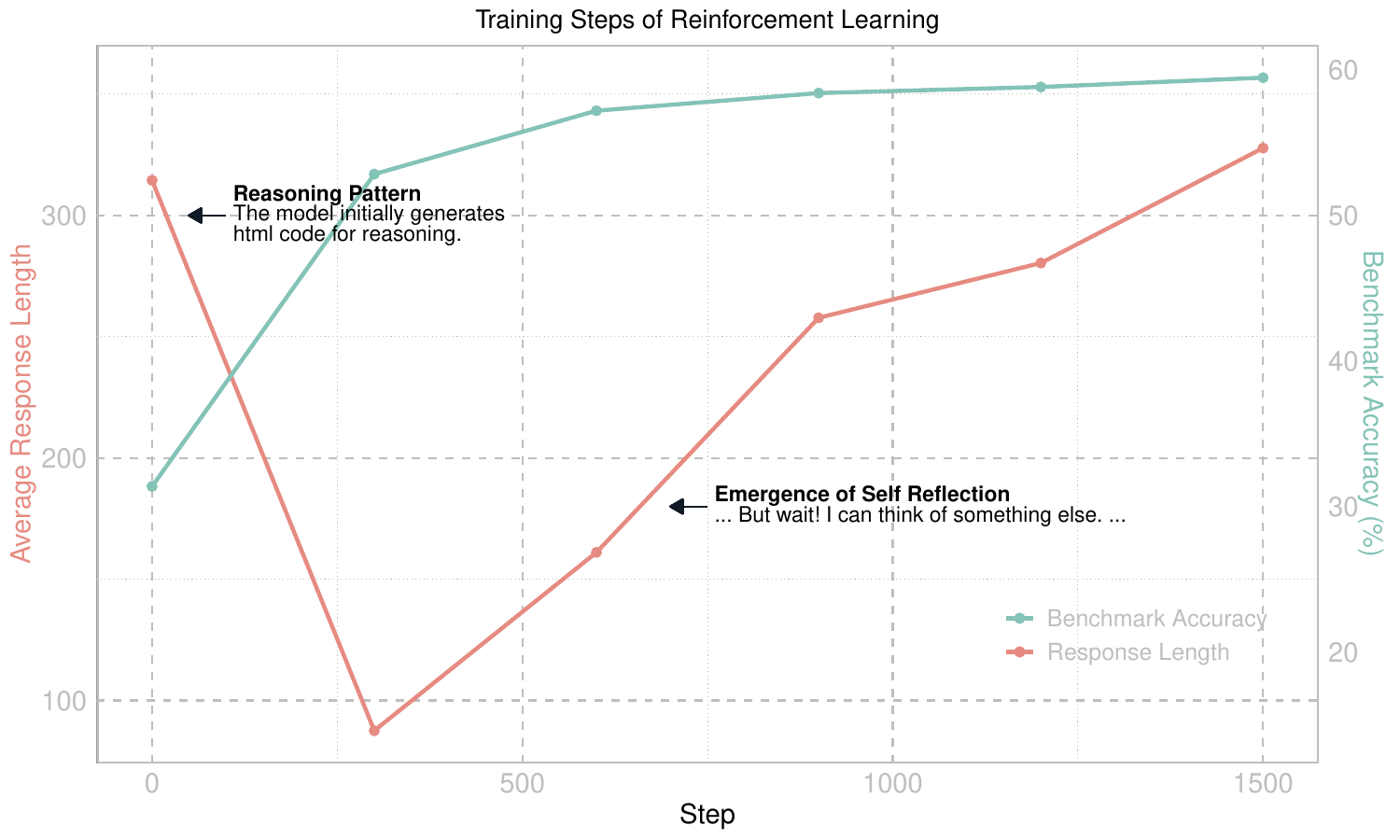}
    \caption{\textbf{The training dynamics of VisualThinker-R1-Zero on Qwen2-VL-2B base model}. Benchmark accuracy is measured on CV-Bench, and the average response length is calculated from rollouts on SAT training samples. Initially, we observed a drop in length because the base model tended to generate HTML code. This behavior was quickly suppressed by RL, leading the model to adopt a more appropriate output format and a regular increase in response length. Afterwards, we observed a multimodal ‘aha moment’—the emergence of self-reflection in models’ response, as described in the DeepSeek-R1 paper, followed by a consistent positive correlation between response length and benchmark accuracy.}
    \label{fig:acc_vs_length}
\end{figure}
% Large language models (LLMs) have shown impressive capabilities across a range of tasks, yet sophisticated reasoning remains challenging~\cite{kumar2025llm, maleki2024ai}. Recent advances like OpenAI's O1~\cite{openai2024learning} have demonstrated that post-training techniques can significantly enhance models' reasoning abilities beyond their pre-trained capabilities.

Recently, DeepSeek R1~\cite{deepseekai2025deepseekr1incentivizingreasoningcapability} has demonstrated how Reinforcement Learning (RL) with simple rule-based incentives can enable a large language model to build complex reasoning capabilities autonomously. A key finding from this work was the emergence of advanced reasoning patterns without explicit supervision—what the researchers termed the "aha moment," characterized by self-reflection, and spontaneous increase in response length during training as the model learned to explore increasingly sophisticated problem-solving strategies.

Many researchers~\cite{chen2025r1v, zheng2025easyr1, shen2025vlmr1, wang-2025-open-r1-video} have attempted to extend this success to multimodal reasoning, where models process and reason on both visual and textual information. However, these efforts have primarily struggled to reproduce the key characteristics exhibited by DeepSeek R1 mentioned above—specifically, the emergent "aha moment" and increased response length during reasoning. These implementations often failed to demonstrate the autonomous development of sophisticated reasoning strategies observed in DeepSeek R1's training.

In this report, we present \textbf{the first successful replication} of these key characteristics for multimodal reasoning on \textbf{only a non-SFT 2B model}. With both "aha moment" and increased length (Figure~\ref{fig:acc_vs_length}),  our approach demonstrates that direct application of reinforcement learning on non-sft model can induce sophisticated reasoning capabilities even in smaller multimodal models without supervised fine-tuning. We start from the Qwen2-VL-2B~\cite{wang2024qwen2vlenhancingvisionlanguagemodels} base model and directly perform reinforcement learning. Without any SFT, our model achieves \textbf{59.47\% accuracy on CVBench}~\cite{tong2025cambrian}, outperforming the base model by approximately \textbf{\textasciitilde30\%} and exceeding the SFT model by \textbf{\textasciitilde2\%}.

In addition to our replication on non-sft model, we also share our insights and failed attempts in achieving R1-like reasoning using RL with instruct model. We observed that starting from a supervised fine-tuned model often failed to reproduce the observations and findings reported by DeepSeek-R1. Upon investigation on this issue, we found that (1) Despite improved performance,  RL on instruct model leads to \textbf{superficial reasoning} rather than genuine problem-solving strategies, and (2) \textbf{naive length reward} are ineffective at inducing deeper reasoning capabilities.

In summary, our key contributions are as follows:
\begin{itemize}
    \item We are \textbf{the first} team to replicate the \textbf{key characteristics of R1 success} ("aha moment" and increased reasoning length) on multimodal reasoning tasks with a \textbf{non-SFT 2B model}.
    \item We showed that \textbf{vision-centric spatial reasoning} tasks could also benefit from improved reasoning capabilities.
    \item We demonstrated that applying RL on instruction-tuned models leads to superficial reasoning.
    \item We open-sourced our project to facilitate future studies on multimodal reasoning.
\end{itemize}

\section{Related Works}

\subsection{Multimodal Reasoning}
Researchers have demonstrated that LLMs can be post-trained to elicit enhanced reasoning abilities~\cite{openai2024learning}. 
With pre-trained visual
encoders to understand visual content alongside textual data, multimodal LLM's reasoning abilities are investigated and enhancement typically requires sophisticated prompting designs~\cite{hu2025visual} or large amounts of reasoning training data~\cite{thawakar2025llamav, xu2024llava}. The research community is increasingly interested in developing more natural methods to incentivize higher intelligence in models without relying on extensive supervised data or complex prompting techniques.

\subsection{The "Aha Moment" Phenomenon in DeepSeek R1}
A recent breakthrough study, DeepSeek R1~\cite{deepseekai2025deepseekr1incentivizingreasoningcapability}, demonstrated that reinforcement learning can incentivize a model's reasoning abilities without any supervised reasoning data. 
Intriguingly, researchers discovered an "aha moment" when directly applying RL with rule-based reward on mathematical datasets—the model autonomously developed advanced problem-solving strategies, including reflection and self-correction.

We summarize the key characteristics that contributed to DeepSeek R1's success and compare them with our model and other multimodal replications in Table~\ref{tab:aha_comparison}. Specifically, we highlight two emergent phenomena: the "aha moment" and increasing response length. The "aha moment" refers to the model's autonomous development of advanced problem-solving strategies during training, while the increasing response length indicates the model naturally learns to allocate more thinking time for reasoning tasks. It remains questionable whether existing multimodal replications~\cite{FanqingM_R1MultimodalJourney, chen2025r1v, EvolvingLMMsLab_openr1multimodal} without these key characteristics can be considered truly valid implementations of the R1 approach.

\begin{table}[htbp]
\centering
\caption{\textbf{Comparison between DeepSeek R1 and its multimodal replications}}
\label{tab:aha_comparison}
\resizebox{\columnwidth}{!}{
\begin{tabular}{lccccc}
 \hline
\toprule
\textbf{Feature} & \textbf{DeepSeek R1~\cite{deepseekai2025deepseekr1incentivizingreasoningcapability}} & \textbf{VisualThinker R1 Zero (Ours)} & \textbf{R1-V~\cite{chen2025r1v}} & \textbf{R1-Multimodal-Journey~\cite{FanqingM_R1MultimodalJourney}} & \textbf{open-r1-multimodal~\cite{EvolvingLMMsLab_openr1multimodal}} \\
\midrule
Base Model & DeepSeek V3 & Qwen2-VL-2B & Qwen2-VL-2B-Instruct & Qwen2-VL-2B-Instruct & Qwen2-VL-2B/7B-Instruct \\
Modality & Language & Vision + Language & Vision + Language & Vision + Language & Vision + Language \\
Aha Moment & Yes & Yes & No & Yes & No \\
Response Length Dynamics & $\uparrow$ & $\uparrow$ & $\downarrow$ & $\downarrow$ & $\downarrow$ \\
\bottomrule
\end{tabular}
}
\end{table}

\section{VisualThinker R1 Zero}
\label{sec:method}
In this section, we demonstrate VisualThinker-R1-Zero have showed an emergence of “aha moment” by directly applying RL training to the non-SFT base model, leading to a superior performance on vision-centric tasks on non-SFT 2B models.

\paragraph{Base Model} Our method builds on Qwen-2-VL-2B~\cite{wang2024qwen2vlenhancingvisionlanguagemodels} as the base model, applying GRPO~\cite{shao2024deepseekmath} with a tailored chat template and prompting strategy to enhance its reasoning capabilities. We posit that applying GRPO to the base model may be a more efficient and effective way to replicate multimodal R1 reasoning compared to training instruct fine-tuned model. As we will show in Section~\ref{sec:failed_attempts}, the instruction-tuned variant, Qwen-2-VL-2B-Instruct, demands significantly more training resources as a base model yet struggles to replicate the emergent ''aha moment'' observed in DeepSeek R1 with notable failure modes.

\paragraph{Training Recipe} We train the model directly on the SAT~\cite{ray2024sat} dataset to let the base model to explore various spatial reasoning for each question $q$ in the dataset $Q$, we apply the following chat template:

% \begin{figure}
%     \begin{tcolorbox}[title=Prompt Template, label=prompt_template]
%     \textbf A conversation between User and Assistant. The user asks a question about the image, and the Assistant solves it. The assistant first thinks about the reasoning process in the mind and then provides the user with the answer.\nUser: {question} \nAssistant: Let me solve this step by step.\n<think> \newline
%     \newline
%     \textbf{Response:} \newline
%     <think>To determine which object is closer to the traffic cone (highlighted by a red box), we need to analyze the distances between the objects in the image.</think> \newline
%     <answer>motorcycle</answer>
% \end{tcolorbox}
%     \caption{Example response from models applying RL on Instruction-Fintuned Qwen2-VL with frozen vision encoder}
%     \label{fig:meaningless_response_frozen_vision}
% \end{figure}

\begin{tcolorbox}[title=Prompt Template, label=prompt_template]
A conversation between User and Assistant. The user asks a question about the image, and the Assistant solves it. The assistant first thinks about the reasoning process in the mind and then provides the user with the answer.\textbackslash n User: \{QUESTION\} \textbackslash n Assistant: Let me solve this step by step.\textbackslash n <think>
\end{tcolorbox}

% \begin{itemize}
%     \item We used Qwen-2-VL-2B as our base model.
%     \item We will show some failed example of Qwen2-VL-2B-Insruct in Section \ref{sec:failed_attempts}
%     \item We used the following chat template and prompting when applying GRPO on Qwen-2-VL-2B
% \end{itemize}

For each of the question $q$ in the dataset, the model generate an response $o$ with this prompt template and it is then optimized using RL objective.

\paragraph{RL algorithm} Existing repos applying RL on top of fine-tuned visual models failed to replicate DeepSeek r1’s key characteristics. In contrast, we witness \textbf{prolonged reasoning trajectory} and \textbf{“aha moment”} with an \textbf{overlooked approach that directly applies GRPO~\cite{shao2024deepseekmath} to an non-SFT model.} Our findings suggest that this setting is \textbf{the key to} \textbf{TRUE “aha moment”} in multimodal reasoning. Now we briefly review the GRPO algorithm we adopted for RL training.

To ease the burden of training an additional value function approximation model employed by PPO~\cite{schulman2017proximalpolicyoptimizationalgorithms}, GRPO adopted the average reward of sampled response of the policy model as the baseline in computing the advantage. Specifically, given the an input question $q$, we first sample a group of responses $\{o_1, o_2, \cdots, o_G\}$ and compute corresponding rewards $\{r_1, r_2, \cdots, r_G\}$ with the reward model. The advantage is then computed as:
\begin{equation}
    \hat{A}_{i, t} = \widetilde{r}_i = \frac{r_i- {\rm mean}(\mathbf{r})}{{\rm std}(\mathbf{r})}
\end{equation}

The policy model is then optimized by maximizing the following KL objective:

\begin{equation}
\begin{split}
    \mathcal{J}_{GRPO}(\theta) &= \mathbb{E}_{q \sim P(Q), \{o_i\}_{i=1}^G \sim \pi_{\theta_{old}}(O|q)}  
    \Bigg[ \frac{1}{G} \sum_{i=1}^G \frac{1}{|o_i|} \sum_{t=1}^{|o_i|} \Bigg\{ \min \Bigg[
    \frac{\pi_\theta(o_{i,t} | q, o_{i,<t})}{\pi_{\theta_{old}}(o_{i,t} | q, o_{i,<t})} \hat{A}_{i,t}, \\
    & \quad \text{clip} \left( \frac{\pi_\theta(o_{i,t} | q, o_{i,<t})}{\pi_{\theta_{old}}(o_{i,t} | q, o_{i,<t})}, 1 - \epsilon, 1 + \epsilon \right)  \hat{A}_{i,t}  
    \Bigg] - \beta \mathbb{D}_{KL}\left[\pi_{\theta} || \pi_{ref}\right] \Bigg\} \Bigg]
\end{split}
\label{eq:GRPO-obj}
\end{equation}

where $\pi_{\theta}$ and $\pi_{old}$ are the current and old policy, and $\epsilon$ and $\beta$ are hyper-parameters introduced in PPO.

\paragraph{Reward Modeling} Following DeepSeek-R1, our RL approach remains elegant, avoiding the use of reward models~\cite{ouyang2022traininglanguagemodelsfollow} or Monte Carlo tree search (MCTS)-like techniques~\cite{zhang2024restmctsllmselftrainingprocess}. Specifically, we employ a rule-based reward function that evaluates responses based on their format and correctness:

\begin{itemize}
    \item If the response provides a final answer and is correct, the model receives an accuracy reward of +1.
    \item If the response encloses its thinking in \texttt{<think></think>} and the final answer in \texttt{<answer></answer>} tags, the model receives a format reward of +1.
    \item otherwise, the model receives 0 reward.
\end{itemize}

Our implementation is based on DeepSeek R1’s report. Initial experiments suggest that this reward function helps the policy model quickly converge towards generating responses in the desired format.

% \subsection{Base Model and Prompt Template}

% \subsection{GRPO}

% \subsection{Reward Modelling}

\section{Experiments}
% \subsection{Task}
In this study, we demonstrate that small non-sft model could solve vision-centric spatial reasoning task with our method, a type of benchmark often pose challenges to even larger models. We train our models on SAT~\cite{ray2024sat}, a VQA dataset comprising 218k question-answer pairs synthesized using a photo-realistic physics engine to enhance spatial intelligence. Our training focuses on the static subset, which includes questions on relative spatial relationships, relative depth, and object counting.
\subsection{Evaluation}
To test the generalization of our method, we evaluate on CVBench~\cite{tong2025cambrian}, a realistic vision-centric benchmark consists of 2,638 examples repurposed from standard vision datasets on fundamental 2D and 3D reasoning tasks. CVBench formulates natural language questions to assess spatial relationships, object counting, depth ordering, and relative distance. This setup allows us to systematically examine how well our recipe could improve spatial reasoning capability.

\subsection{Implementation Details}
We conduct all our experiments using four NVIDIA H100 GPUs (80GB each), setting the batch size to 1 per device. The model is trained for 1500 steps with a learning rate of $1 \times 10^{-6}$ and a temperature of 1.0. We found long response length is a must for observing increasing response length during training, so we set the maximum response length as 700. During GRPO optimization, we sample 8 responses per step and apply a KL coefficient of 0.04.

\begin{table}[h]
    \centering
        \caption{\textbf{Hyper-parameters of VisualThinker R1 zero GRPO training.}}
        \resizebox{0.5\columnwidth}{!}{
    \begin{tabular}{l c}
             \hline
            \toprule
        \textbf{Setting} & \textbf{Value} \\
        \hline
        Batch Size per Device & 1 \\
        Gradient Accumulation Steps & 2 \\ 
        Training Steps & 1500 \\
        Learning Rate & $1 \times 10^{-6}$ \\
        Temperature & 1.0 \\
        Maximum Response Length & 700 \\
        Number of Responses per GRPO Step & 8 \\
        KL Coefficient & 0.04 \\
        \hline
    \end{tabular}}
    \label{tab:exp_settings}
\end{table}
To demonstrate the superior effectiveness of directly applying GRPO on base model, we compare our method with the non-sft pre-trained mode as the baseline, and compare our method against the base model SFT on the same SAT dataset.
% Key hyperparameters
% Steps, KL Coefficient, Max Response Length, Max Pixels, Num of generations

\subsection{Main Results}
\begin{figure}[H]
    \centering
    \includegraphics[width=0.9\textwidth]{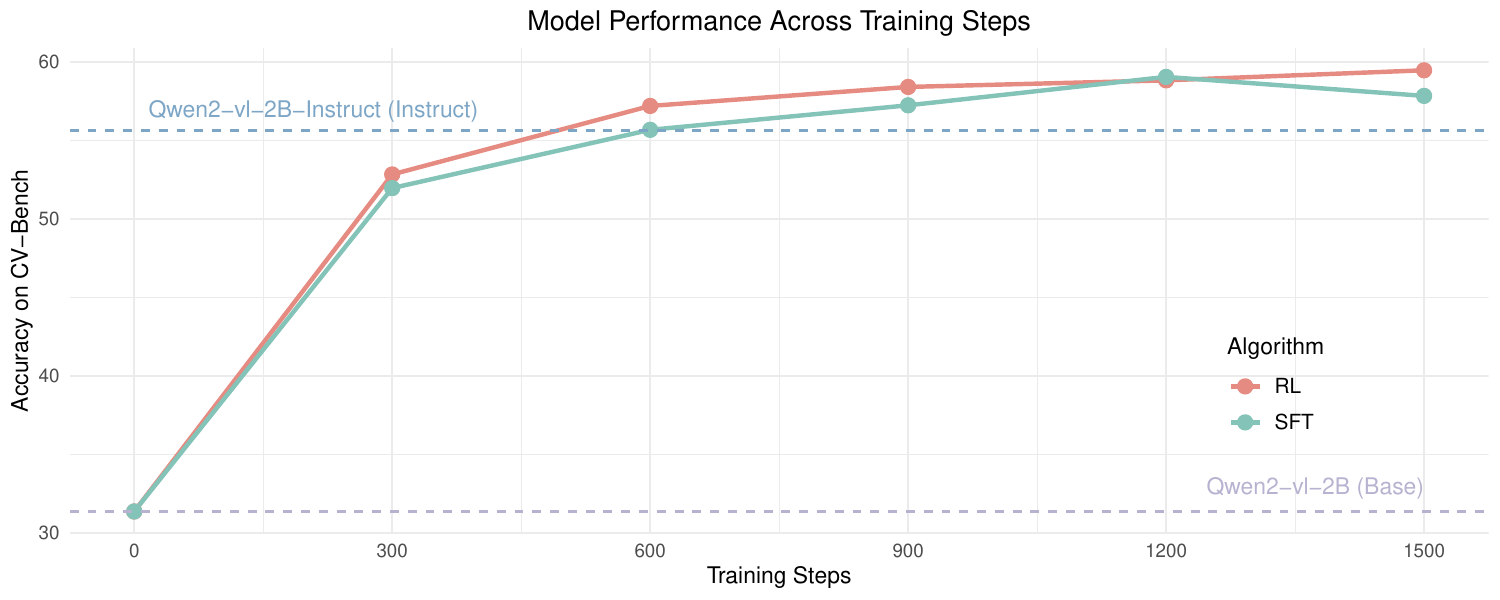}
    \caption{\textbf{Comparison between RL and SFT training.} Our method achieves a significant improvement over the base model and the instruction fine-tuned model. Specifically, Qwen2-VL-2B + R1 outperforms Qwen2-VL-2B (base model) by approximately \textasciitilde30\%, Qwen2-VL-2B-Instruct (instruction fine-tuned model) by \textasciitilde5\%, and Qwen2-VL-2B SFT (base model + SFT) by \textasciitilde2\%.}
    \label{fig:RL_vs_SFT}
\end{figure}

\paragraph{Benchmark Results} In the experiments, we fine-tuned the Qwen2-VL-2B non-sft model and evaluated its performance on CV-Bench. During training, we can observe the model autonomously develop increasing response length and the performance comes with it in Figure~\ref{fig:acc_vs_length} on CV-Bench. We can also observe that directly applying RL on base model has superior performance compared to SFT method in Figure~\ref{fig:RL_vs_SFT}.

In addition, we tested our method on various spatial reasoning dataset including BLINK~\cite{fu2024blinkmultimodallargelanguage}\footnote{Following SAT~\cite{ray2024sat} evaluation, we use two spatial splits of BLINK - Multiview reasoning, Relative Depth, and Spatial Relations.} and VSR~\cite{liu2023visualspatialreasoning}, illustrated in Table~\ref{tab:main_result}. Our method demonstrated improved performance over the Qwen2-VL-2B (base) by \textasciitilde30\%, and the Qwen2-VL-2B SFT(base + SFT) by \textasciitilde2\% on CV-Bench. We also achieve superior performance on the BLINK and VSR benchmark, with our method achieves around 27\% advantage comparing against the model trained with SFT. This suggests that visual reasoning could significantly benefit from R1-Zero training, demonstrating more scalable training through RL’s exploration of diverse reasoning.

\begin{table}[H]
    \centering
    \caption{\textbf{Results on vision-centric benchmarks}. Table shows RL training on base model has overall better performance over SFT training and the base model.}
    \resizebox{\columnwidth}{!}{
        \begin{tabular}{lcccccccccc}
             \hline
            \toprule
             \multicolumn{1}{c}{Model} & \multicolumn{5}{c}{CV-Bench} & \multicolumn{3}{c}{BLINK} & \multicolumn{1}{c}{VSR}\\
             \midrule
              & Count Acc(\%) & Relation Acc(\%) & Depth Acc(\%) & Distance Acc(\%) & Total Acc(\%) & Relative Depth Acc(\%) & Spatial Relation Acc(\%) & Average Acc(\%) & Average Acc(\%) & \\
              \hline
             \textbf{Qwen2-VL-2B} & 54.69 & 22.46 & 0.16 & 31.66 & 31.38 & 13.70 & 0.69 & 6.74 & 0.0 & \\
               \textbf{Qwen2-VL-2B + SFT} & 60.02 & 68.92 & 55.00 & 45.83 & 57.84 & 58.06 & 47.55 & 52.43 & 35.80 & \\
              \textbf{Qwen2-VL-2B + GRPO (Ours)} & 59.64 & 66.76 & 54.16 & 56.66 & 59.47 & 50.80 & 55.94 & 53.18 & 62.32 & \\
             \hline
             % & Qwen2-VL-2B + GRPO \\
             % & 
        \end{tabular}
    }
    \label{tab:main_result}
\end{table}

\paragraph{Multimodal Aha Moment} A particularly intriguing phenomenon observed during the training of DeepSeek-R1-Zero is the occurrence of an “aha moment”:

\begin{verbatim}
...
**Wait, wait. Wait. That’s an aha moment I can flag here.**
Let’s reevaluate this step-by-step to identify if the correct sum can be...
We started with the equation:
...
\end{verbatim}
This aha moment indicates DeepSeek-R1-Zero spontaneously builds a reasoning strategy rethinking its initial approach for improved reasoning capability.

Beyond strong performance gain, we also observed our model exhibit very interesting "aha moment" aligned with the finding in DeepSeek R1: during training, the model spontaneously revisit its previous judgment and explore alternative options:

\begin{verbatim}
...
Therefore, dark brown wooden bed with white blanket is not above 
the doorway. 
**But wait! I can think of something else.** <think>
...
\end{verbatim}

% In Figure~\ref{fig:keyword_dist}, we illustrate the emergence of complex reasoning behavior measured by specific keywords. We can observe that the model eventually learn to re-check, verify its previous step, think about alternatives, and catch the aha moment around step 800.

% \begin{figure}[H]
%     \centering
%     \includegraphics[width=0.8\textwidth]{images/keyword_plot_wide.pdf}
%     \caption{\textbf{Keyword Distribution}: The model spontaneously develop complex reasoning behavior such as re-checking its previous steps, explore alternative solutions, and catch the aha moment around step 800.}
%     \label{fig:keyword_dist}
% \end{figure}
% Distribution of keywords
% Keywords related to reward

\section{Challenges of Applying RL to Supervised Fine-Tuned Models}
\label{sec:failed_attempts}

Starting from scratch with a non-SFT multimodal model, we have demonstrated that RL empowers VisualThinker-R1-Zero to attain robust reasoning capabilities without any supervised fine-tuning data. In addition to this replication, one might be inclined to apply RL directly to supervised fine-tuned models given its stronger instruction following capability.

In this section, we share our failed cases on applying GRPO algorithm described in Section~\ref{sec:method} to the Qwen2-VL-2B-Instruct model (a supervised fine-tuned Qwen VL model) to provide insights that may benefit future research. However, note that this does not imply that these approaches are incapable of building effective visual reasoning models.
Specifically, we found that starting from a supervised fine-tuned model exposes problems of trival reasoning patterns. We then demonstrate our findings in (1) investigating the reason and implication of this phenomenon and (2) attempting to incentivize sophisticated reasoning. 
% Our findings suggest two primary challenges of applying RL on supervised fine-tuned models: (1). RL leads to \textbf{superficial reasoning patterns}. (2). \textbf{Direct reasoning incentives} are ineffective.

\subsection{Emergence of Trivial Reasoning Patterns}

\begin{figure}
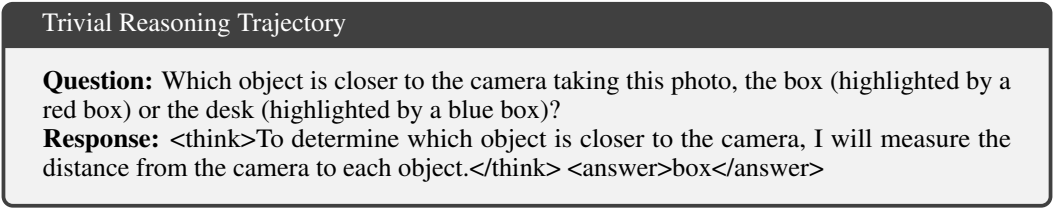

    \begin{tcolorbox}[title=Trivial Reasoning Trajectory, label=meaningless_response]
    \textbf{Question:} Which object is closer to the camera taking this photo, the box (highlighted by a red box) or the desk (highlighted by a blue box)? \newline
    \textbf{Response:} <think>To determine which object is closer to the camera, I will measure the distance from the camera to each object.</think> <answer>box</answer>
\end{tcolorbox}
    \caption{\textbf{Example response of applying RL to supervised fine-tuned models.}}
    \label{fig:meaningless_response}
\end{figure}

Applying RL on an SFT model does improve the model's performance on CVBench (Table~\ref{tab:meaningless_reasoning}). However, it is questionable whether this method incentivizes the models to achieve higher intelligence, as Figure~\ref{fig:meaningless_response} demonstrates the degeneration of model responses into meaningless or trivial reasoning patterns.
The observed reasoning trajectory follows a structure: a trivial and generic strategy within \texttt{<think></think>} tags followed by a question-specific answer between \texttt{<answer></answer>} tags. One hypothesis is that the model enhances its performance during RL training without necessarily developing genuine reasoning capabilities.

% \paragraph{Impact of Reasoning} To further verify the necessity of evolving reasoning capabilities, we compared different prompting approaches. Rather than prompting models to generate intermediate reasoning (Figure~\ref{fig:different_prompting_on_instruct}), we instructed models to respond directly to questions without reasoning steps. As demonstrated in Table~\ref{tab:meaningless_reasoning}, when using direct-answer-only prompting, RL achieved approximately 9\% better performance compared to prompting that required both reasoning and answers. This suggests that the reasoning behavior may be unnecessary or even detrimental to performance improvements when applying RL on instruct models.

% \begin{figure}
%     \begin{tcolorbox}[title=Prompting with both \texttt{<reason>} and \texttt{<answer>}, label=prompting_reason_answer]
%     \{QUESTION\} Output the thinking process in \texttt{<think> </think>} and final answer in \texttt{<answer> </answer>} tags.
% \end{tcolorbox}
%     \begin{tcolorbox}[title=Prompting with only \texttt{<answer>}, label=meaningless_response]
%     \{QUESTION\} Output the final answer in \texttt{<answer> </answer> tags}.
% \end{tcolorbox}
%     \caption{\textbf{Different prompting strategy} to invesitgate the impact of reasoning in RL training on instruct model.}
%     \label{fig:different_prompting_on_instruct}
% \end{figure}

\subsection{Preliminary Investigation of Trivial Reasoning Trajectories}

% \paragraph{Component-wise Contribution Analysis} 
Given the trivial reasoning trajectories, a nature question is whether performance improvements occur through enhancement of the vision encoder during training. We hypothesized that when we freeze the vision encoder during RL, the model may focus more on developing sophisticated reasoning strategies; conversely, when we freeze the LLM during RL, the performance would be the same level as the full fine-tuning setting. Surprisingly, results in Table~\ref{tab:frozen-parts} demonstrate that both approaches achieve greater improvement than the vanilla implementation yet they still generate short and trivial responses. This observation suggests the complexity of training dynamics of RL on multimodal models and further analysis is needed to better understand the phenomenon.

% To trace back to where does the improvement occur, we conducted ablation studies by freezing different components of the multimodal LLM—specifically, the vision encoder or the language model—to determine whether the improved performance stems from enhanced perception or language capabilities. As shown in Table~\ref{tab:frozen-parts}, although training multimodal LLMs while freezing the visual encoder improves performance, we can also achieve comparable improvement by only training on vision encoder.
% However, both experimental settings—freezing the LLM and freezing the vision encoder—deliver short and meaningless responses like the one shown above (see for concrete examples). 
% This finding suggests that more fine-grained studies are needed to understand exactly whether reasoning ability improvements occur during RL training on instruction-tuned models.

\begin{table}[h]
    \centering
    \caption{\textbf{Evaluation of applying RL to instruct models.} Prompting with only \texttt{<answer>} achieves superior performance in the context of applying RL on instruct models.}
    \resizebox{\columnwidth}{!}{
    \begin{tabular}{lcccccc}
         \hline
        \toprule
        & Prompting Strategy & Total Acc (\%) & Count Acc (\%) & Relation Acc (\%) & Depth Acc (\%) & Distance Acc (\%) \\
        \midrule
        Instruct Model & - & 55.64 & 45.43 & 68.92 & 58.66 & 51.66 \\
        % Instruct Model + RL & without reason & 75.70 & 68.02 & 82.00 & 78.66 & 76.00 \\
        Instruct Model + RL & with reason & 66.03 & 69.54 & 61.84 & 66.50 & 65.50 \\
        \bottomrule
    \end{tabular}
    }
    \label{tab:meaningless_reasoning}
\end{table}

\begin{table}[hbt!]
    \centering
    \caption{\textbf{Evaluation of applying RL to instruct models with different freezing components.} Either freezing language or vision components of instruct model can both leads to performance improvement.}
    \resizebox{\columnwidth}{!}{
    \begin{tabular}{lcccccc}
         \hline
        \toprule
        & Freezing Components & Total Acc (\%) & Count Acc (\%) & Relation Acc (\%) & Depth Acc (\%) & Distance Acc (\%) \\
        \midrule
        Instruct model + RL &- & 63.57 & 61.68 & 71.08 & 61.50 & 60.00 \\
        Instruct model + RL & vision encoder & 68.34 & 64.21 & 78.00 & 65.67 & 66.00 \\
        Instruct model + RL & LLM & 65.35 & 62.05 & 72.46 & 65.17 & 62.17 \\
        \bottomrule
    \end{tabular}
    }
    \label{tab:frozen-parts}
\end{table}

\begin{figure}[hbt!]
    \centering
    \includegraphics[width=0.8\linewidth]{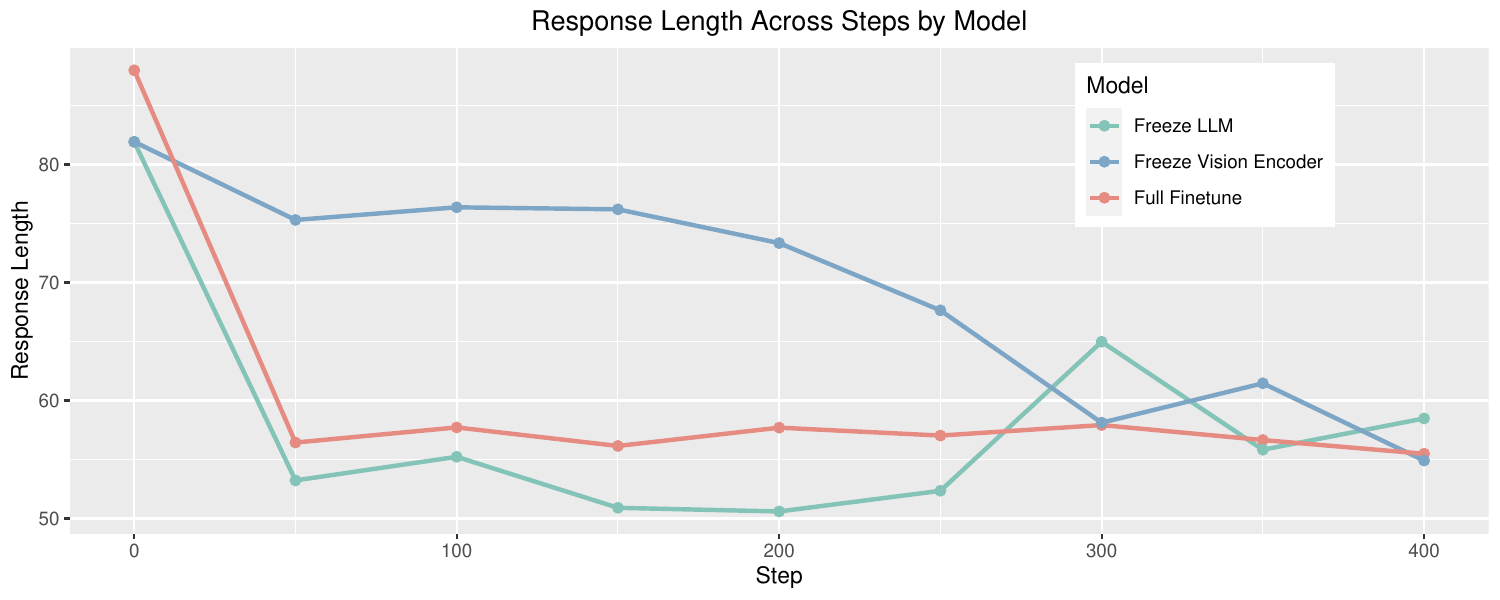}
    \caption{\textbf{Response length across training steps for different fine-tuning settings during RL.} The x-axis represents training steps, while the y-axis shows the response length. Models with different fine-tuning configurations are compared: Freeze LLM (green), Freeze Vision Encoder (blue), and Full Finetune (red). The response length drops significantly in the early training phase and stabilizes over time. However, despite improved accuracy, all three RL-based fine-tuning on Instruct Model does not necessarily enhance reasoning capabilities, as responses tend to remain short and trivial}
    \label{fig:frozen-parts}
\end{figure}

% \subsection{Ineffectiveness of Direct Reasoning Incentives}
\subsection{Failed Attempts with Length Reward}
The key characteristics of DeepSeek R1's success is its increased response length during training. 
After observing our failure to replicate DeepSeek R1's reasoning capabilities when starting from instruction-tuned models, we investigated whether we could encourage more sophisticated reasoning patterns by directly rewarding longer response.
% We implemented a length-based reward mechanism to stimulate more detailed reasoning outputs.

% \paragraph{Reflection-Orientated Prompting} To incentivize meaningful reasoning development during RL, we employ a simple prompting strategy designed to guide the model toward reflective thinking. Specifically, we use the following system prompt and prompt suffix in Figure~\ref{fig:rething_prompting}.
% However, [experiment results]

% \paragraph{Length-Based Rewards} 

\begin{figure}[hbt!] % 
    \centering
    \includegraphics[width=0.8\textwidth]{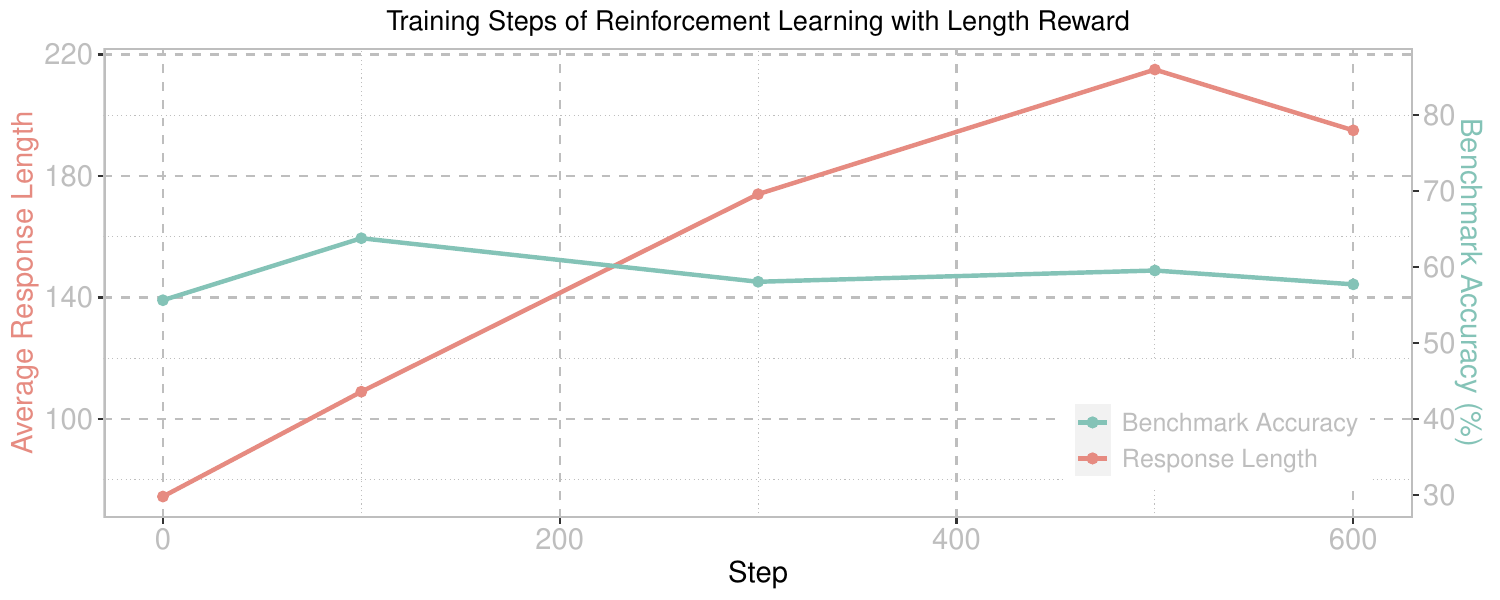}
    \caption{\textbf{Model performance during training with length-rewarded RL on instruction ginetuned models.}  Despite implementing length rewards to encourage longer responses, model performance shows no improvement in reasoning quality as response length increases. While average response length (red line) grows substantially, benchmark accuracy (green line) remains relatively stable, suggesting that longer responses do not necessarily translate to enhanced reasoning capabilities.}
    \label{fig:length_reward}
\end{figure}

\begin{figure}[bt!]
    \begin{tcolorbox}[title=Length Reward Hacking Example, label=lengthreward]
    \textbf{Question:} Considering the relative positions of the giraffe (annotated by the red box) and the person in the image provided, where is the giraffe (annotated by the red box) located with respect to the person? \newline
    \textbf{Model Response:} <think> - The giraffe (annotated by the red box) is located in the foreground of the image.  - The person is in the background, standing behind the giraffe.  - The giraffe is closer to the viewer than the person.  - The person is not directly above the giraffe.  - The person is not directly below the giraffe.  - The giraffe is not above the person.  - The giraffe is not below the person. - The person is not in front of the giraffe. - ... $[$Repeats until it reaches max\_new\_tokens limit$]$  ... </think>
    \newline
\end{tcolorbox}
    \caption{\textbf{Response example of model hacking naive length reward}. Model hacks reward by repeating trivial response.}
    \label{fig:length_reward_hacking}
\end{figure}

Thus, we implemented a length-based reward mechanism: in addition to the vanilla accuracy and format rewards described in Section~\ref{sec:method}, we added an auxiliary reward of $+0.001$ for each additional token generated. 
However, as shown in Figure~\ref{fig:length_reward}, naively rewarding lengthy responses does not improve model performance and often leads to reward hacking behaviors, with models generating extremely long yet meaningless content as illustrated in Figure~\ref{fig:length_reward_hacking}.
This finding suggests that the emergent reasoning capabilities observed in DeepSeek R1 cannot be easily achieved through enforcing long response length during the training of instruction-tuned models. 
% Our experiments with both reflection-oriented prompting and length-based rewards failed to induce real characteristics of DeepSeek r1 in instruction-tuned models. The reflection prompting approach did not lead to meaningful self-correction or deeper analysis, while the length-based rewards caused models to exploit the reward mechanism by generating repetitive, content-poor responses rather than substantive reasoning. 
% This finding suggests that the emergent reasoning capabilities observed in DeepSeek R1 cannot be easily reproduced through direct incentive mechanisms alone when starting from instruction-tuned models. 
% (Instead, the development of sophisticated reasoning patterns likely requires training from less constrained base models that can discover novel problem-solving strategies without being limited by patterns established during supervised fine-tuning.) Considering

% \begin{figure}
%     \begin{tcolorbox}[title=System Prompt, label=systemprompt]
%     You are a helpful assistant capable of complex reasoning and reflection. 
%     \end{tcolorbox}
%     \begin{tcolorbox}[title=Chat Template, label=chattemplate]
%     \{QUESTION\} Output the thinking process in \texttt{<think></think>} and final answer in \texttt{<answer></answer>} tags. If you detect that you made a mistake in your reasoning at any point, correct yourself inside \texttt{<reflection></reflection>} tags.
%     \end{tcolorbox}
%     \caption{System prompts and chat template employed to incentivize models' meaningful reasoning with reflection.}
%     \label{fig:rething_prompting}
% \end{figure}

\section{Conclusion}
This paper presents VisualThinker R1 Zero, the first successful multimodal replication of DeepSeek R1's emergent reasoning characteristics. By applying reinforcement learning directly to a non-fine-tuned Qwen2-VL-2B model, we observed both the "aha moment" and increased response length during training—key indicators of autonomous reasoning development. Empirically, our approach achieved 59.47\% accuracy on CVBench, outperforming both base and instruction-tuned models without any supervised fine-tuning. We also share our insights and failed attempts in achieving R1-like reasoning using RL with instruct model: applying RL to supervised fine-tuned models leads to trivial reasoning trajectories rather than genuine problem-solving strategies. 

This report is a work in progress for presenting our preliminary findings, we plan to release further updates with deeper investigations and insights to continuously expand and refine this report with our ongoing exploration of the technical roadmap for realizing R1-like multimodal reasoning.

% This report serves as an initial replication of R1 on multimodal reasoning. We aim to extend and update this report with ongoing research on investigation and analysis of this replication.

% \begin{ack}
% Use unnumbered first level headings for the acknowledgments. All acknowledgments
% More information about this disclosure can be found at: \url{https://neurips.cc/Conferences/2024/PaperInformation/FundingDisclosure}.

% Do {\bf not} include this section in the anonymized submission, only in the final paper. You can use the \texttt{ack} environment provided in the style file to automatically hide this section in the anonymized submission.
% \end{ack}

\bibliographystyle{plain} % You can use different styles like 'plain', 'unsrt', 'alpha', etc.
\bibliography{references}

\end{document}